\documentclass[a4paper]{article}

\usepackage[english]{babel}
\usepackage[utf8x]{inputenc}
\usepackage[T1]{fontenc}

\usepackage[a4paper,top=3cm,bottom=2cm,left=3cm,right=3cm,marginparwidth=1.75cm]{geometry}
\usepackage{amsthm}%
\usepackage{graphicx}%
\usepackage{multirow}%
\usepackage{amsmath,amssymb,amsfonts}%
\usepackage{mathrsfs}%
\usepackage[title]{appendix}%
\usepackage{xcolor}%
\usepackage{textcomp}%
\usepackage{manyfoot}%
\usepackage{booktabs}%
\usepackage{algorithm}%
\usepackage{algorithmicx}%
\usepackage{algpseudocode}%
\usepackage{listings}%
\usepackage{ulem}%
\usepackage{array}%
\usepackage{multirow}%
\usepackage{booktabs}%
\usepackage{natbib}
\usepackage{hyperref}%
\usepackage{comment}%
\usepackage{titling}%
\usepackage{url}%
\usepackage{array} %
\usepackage{tablefootnote} %
\usepackage{tabularx}%
\usepackage{pdflscape}%
\usepackage{float}%

\usepackage{tikz} 
\usetikzlibrary{bayesnet}
\usetikzlibrary{arrows}

\newtheorem{Definition}{Definition}

\newcommand{\keywords}[1]{%
  \vspace{2ex}\noindent\textbf{Keywords:} #1
}

\newcommand{\ind}{\mbox{$\perp\!\!\!\perp \,$}}

\definecolor{mygreen}{RGB}{203, 223, 219}

\begin{document}
\title{Uncertainty-Aware Fairness-Adaptive Classification Trees}

\author{
  Anna Gottard\textsuperscript{1}
  \and
  Vanessa Verrina\textsuperscript{2} \and
  Sabrina Giordano\textsuperscript{2}
}

\date{
  \textsuperscript{1}Department of Statistics, Computer Science, Applications ``G. Parenti'',\\ University of Florence, Italy\\
  \textsuperscript{2}Department of Economics, Statistics and Finance ``Giovanni Anania'', \\University of Calabria, Italy\\[2ex]
}
\maketitle

\begin{abstract}
In an era where artificial intelligence and machine learning algorithms increasingly impact human life, it is crucial to develop models that account for potential discrimination in their predictions. This paper tackles this problem by introducing a new classification tree algorithm using a novel splitting criterion that incorporates fairness adjustments into the tree-building process. The proposed method integrates a fairness-aware impurity measure that balances predictive accuracy with fairness across protected groups. By ensuring that each splitting node considers both the gain in classification error and the fairness, our  algorithm encourages splits that mitigate discrimination. Importantly, in penalizing unfair splits, we account for the uncertainty in the fairness metric by utilizing its confidence interval instead of relying on its point estimate.  Experimental results on benchmark and synthetic datasets illustrate that our method effectively reduces discriminatory predictions compared to traditional classification trees, without significant loss in overall accuracy.
\end{abstract}


\section{Introduction}\label{sec:intro}

The rapid advancements in machine learning (ML) and artificial intelligence (AI) have brought substantial changes across various sectors, including healthcare, finance, education, and criminal justice. See, among others, 
 \citep{sarker2021machine},  \citep{azar2013decision}, \citep{giudici2023safe}, \citep{cortez2008student},  \citep{dressel2018accuracy}.
These ML procedures promise to improve decision-making processes, increase efficiency, and uncover new insights from large amounts of data. However, their deployment has also raised significant ethical and social concerns, particularly regarding fairness, accountability, and transparency \citep{valera2021discrimination}. 

Fairness in algorithms is a multifaceted and complex issue,  involving the prevention of perpetuating or worsening existing biases. As algorithms are trained on data, they often inherit and amplify biases present in the sample, leading to discriminatory outcomes that  affect  vulnerable groups. For example, biased algorithms in hiring practices can disadvantage applicants based on gender or ethnicity \citep{chen2023ethics}, while biased predictive policing models can lead to over-policing of minority communities \citep{engel2024code,angwin2022machine}.
In response to these concerns, machine learning researchers, in collaboration with policymakers, social scientists, and other stakeholders, have increasingly focused on the topic of fairness in ML, particularly on designing fair classification and regression algorithms  \citep{barocas2016big, EUcommission2020}.

Bias in ML algorithms can stem from their goal of minimizing overall prediction errors, which may unintentionally favor majority groups and worsen systemic inequities. Additionally, excluding sensitive attributes like race, gender, and age to prevent discrimination often leads models to rely on proxy attributes indirectly linked to these factors, inadvertently perpetuating bias. These issues highlight the complex nature of fairness in ML systems and underscore the need for comprehensive bias mitigation strategies. Special attention is needed for the trade-off between accuracy and fairness, as prioritizing fairness might sometimes reduce accuracy \citep{kleinberg2017inherent}.

There are three primary approaches in ML to enhance fairness. The first, known as pre-processing, involves altering the training data to reduce or remove bias before model training. The second approach, post-processing, adjusts model predictions after training to meet fairness criteria, by recalibrating outputs or modifying decision thresholds. The third, in-processing,  incorporates fairness directly into the learning algorithm during training. This can be achieved by adding fairness constraints or penalty terms to the objective function to mitigate undesired bias, as discussed in \cite{pessach2022review}. 

In this paper, we adopt an in-processing approach to introduce an uncertainty-aware classification tree algorithm designed to mitigate discrimination. 
Tree-based algorithms, particularly the Classification and Regression Trees (CART) algorithm \citep{breiman1984classification}, are popular in ML due to several advantages. They provide a simple visual representation of how predictions are made when using a single tree like CART, enhancing transparency. They can handle mixed data types and missing values, and they can employ any computationally tractable splitting criterion, making them adaptable to specific requirements.
However, a drawback is that they tend towards unfair predictions due to the greedy search of the tree structure, which can lead to an over-reliance on the background variables \citep{gottard2020note}. Therefore, there is a need to address this issue and make classification trees and tree-based algorithms fairer.
In this direction, \cite{kamiran2010discrimination} proposes a modified version of classification trees involving both in-processing and post-processing approaches. For selecting tree splits, they introduced an Information Gain measure that combines gain in fairness, seen as entropy for the sensitive attribute,  with gain in accuracy.  
Additionally, they implement a post-processing procedure to relabel the predictions in the terminal nodes. Instead of using the ordinary majority class criterion, they select predictions that reduce discrimination while maintaining minimal loss in accuracy. 
On the other hand, \cite{zhang2019faht} provide a fair version of the Hoeffding tree classifier proposed by \cite{domingos2000mining} for streaming data. They propose to modify the split gain as the product of the ordinary information gain times a fairness gain based on statistical parity (see Section \ref{sec:metrics} for a formal definition).
Beyond the CART algorithm, \cite{aghaei2019learning}, building on the work of \cite{bertsimas2017optimal}, propose using integer optimization to design fair decision trees. Their method adds a fairness regularizer to the loss function, penalizing discrimination in terms of disparate treatment or disparate impact metrics, both in the case of classification and regression. \cite{castelnovo2022fftree} introduce a flexible fairness-aware decision tree that allows users to select multiple fairness metrics and incorporate several sensitive attributes simultaneously. The desired fairness level needs to be pre-specified as an input parameter. 
Most recently, \cite{pereira2024fair} propose a new class of classification trees fair splitting criterion that reaches accuracy adopting a loss function based on the threshold-independent ROC-AUC classification error and fairness using the threshold-independent fairness measure termed strong statistical parity, proposed by \cite{jiang2020wasserstein}. 

We propose an Uncertainty-Aware Adaptive Fair classification tree algorithm, which is a modified version of the CART algorithm. In our proposal, the Information Gain is modified to include a penalty factor that accounts for the uncertainty in the fairness metric. 
Specifically, when evaluating potential splits during the tree-building process, we compute a confidence interval of the fairness metric associated with each split. If this confidence interval does not include zero, it suggests that the split contributes significantly to discrimination. We then penalize these splits with a factor that adapts to the degree of discrimination suggested by the confidence interval.
This adaptive penalty works by decreasing the Information Gain for splits that are more strongly discriminatory, effectively discouraging the algorithm from selecting them. By incorporating this penalty, the tree construction process naturally favours splits that are both accurate and fair, without requiring a separate post-processing step.
The penalization factor can be tuned to balance the well-known trade-off between accuracy and fairness, allowing users to adjust  the degree of fairness mitigation to their specific needs. As a result, the final classification tree is better equipped to make equitable predictions, reducing discrimination against protected groups while maintaining good predictive performance.
In this paper, we focus on the case where both the response variable and the sensitive attribute are binary. However, the method can be extended to multi-category cases by appropriately modifying the definition of the confidence intervals.

The remainder of the paper is organized as follows. In Section \ref{sec:notions}, we briefly introduce the Statistical Parity measure and its confidence interval. Section \ref{sec:fair_cart}
proposes the adaptive fair algorithm to build a classification tree with mitigated statistical disparity. Section \ref{sec:exper} presents experimental results on benchmark and synthetic datasets.
Finally, we conclude the paper in Section \ref{sec:conclusions}.

\section{Preliminaries}\label{sec:notions}

Let us consider a binary supervised classification problem with a random sample $\{\mathbf{X_i}, S_i, Y_i\}_{i=1}^{n}$, where $\mathbf{X_i}$ is the feature vector with \textit{p} predictors of the $i$th unit, $S_i$ is the sensitive attribute (e.g., race or gender) and $Y_i$ is the binary response variable. In particular, we assume that $S_i=1$ for the privileged group and $S_i = 0$ for the unprivileged group. Moreover, $\widehat{Y}_i$ is the predicted value for $Y_i$ with $1$ denoting the positive outcome. 

\subsection{A fairness metric and its uncertainty}\label{sec:metrics}
Fairness in automated decision-making systems is a multifaceted and context-dependent concept \citep[see, among others,][]{pedreschi2018blackbox}. Fairness notions in the literature can be broadly categorized into definitions based on statistical parity across different groups (e.g., gender, age), prevention of disparate treatment among similar individuals, and the use of causality to avoid unfair impacts \citep[see, among others,][]{castelnovo2022clarification, imai2023principal}.

Group fairness, specifically, ensures equitable treatment across sub-groups. Group fairness criteria are typically expressed as conditional independence statements among the relevant variables
in the problem. One key measure of group fairness is the \textit{statistical parity}, which requires that individuals from different groups have similar probabilities of receiving positive outcomes. This helps prevent systematic favoritism or disadvantage. If a group has a significantly lower positive prediction ratio compared to others, the statistical parity is not met. A metric which measures the maximum difference  of positive prediction ratios is used to summarize disparity, with values close to 0 indicating fairness. Some tolerance is typically allowed, with thresholds determining acceptable decisions \citep{pessach2022review}. 

\begin{Definition} The Statistical Parity (SP) is  expressed as:
\begin{equation}
\label{eq:SP}
    \Delta= P\left [\widehat{Y}=1 \mid S=1 \right ]-P\left [\widehat{Y}=1 \mid S = 0 \right ]. 
\end{equation}
where $S$ represents the sensitive variable. When $S = 1 $ is the privileged
group and $S=0$ is the unprivileged group, this measure assumes positive values, provided that $\widehat{Y}=1$ represents a positive output. 
\end{Definition}

Positive values of $\Delta$ indicate that the privileged group has a higher chance of receiving positive outcomes compared to the unprivileged group. Notably, if $\widehat{Y}=1$ corresponds to an acceptance decision (e.g., job offer), achieving fairness would require that acceptance rates be similar across different groups, such as men and women. A smaller value of $\Delta$ reflects more similar acceptance rates between groups, suggesting better fairness in the decision-making process \citep{pessach2022review}.

One limitation of this measure is that even a perfectly accurate classifier may be considered unfair if the underlying proportions of positive outcomes vary significantly between groups. Additionally, aiming for group parity can lead to individuals who are otherwise similar being treated differently based solely on their group membership.
The above fairness measure is typically estimated from sample data and thus subject to sampling variability, which is often overlooked. Specifically, the estimates for $P[\widehat{Y}=1 \mid S=1]$ and $P[\widehat{Y}=1 \mid S=0]$, representing the probabilities of positive predictions in the privileged and unprivileged groups, respectively, are calculated as the sample proportions. To account for sampling uncertainty, confidence intervals (CIs) for these proportions can be constructed.

Let ${n}_{11}$ and ${n}_{10}$ denote the observed frequencies of positive predictions in the privileged and unprivileged groups, with sample sizes $n_1$ and $n_0$. The corresponding estimated proportions are $\widehat{p}_{11}={n}_{11}/n_1$ and $\widehat{p}_{10}={n}_{10}/n_0$. The asymptotic $(1-\alpha)\%$ Wald confidence interval $\text{CI}(\Delta)$ for the Statistical Parity is given by
\begin{equation} \label{eq:CI}
\widehat{p}_{11} - \widehat{p}_{10} \mp z_{\alpha/2} \sqrt{\frac{\widehat{p}_{11}(1-\widehat{p}_{11})}{n_1} + \frac{\widehat{p}_{10}(1-\widehat{p}_{10})}{n_0}},    
\end{equation}
where $z_{\alpha/2}$ represents the upper ${\alpha}/2$ percentile of the standard normal distribution.
While these intervals are easy to compute, more sophisticated methods for constructing confidence intervals for differences,  and also ratios of proportions, particularly in small samples, are available and can be used in the proposed algorithm. For a comprehensive overview of these confidence intervals, see \cite{fagerland2017statistical}, and, for a review of score-test-based CIs, see \cite{agresti2022review}.

\section{The proposed algorithm}\label{sec:fair_cart}

In this section, we propose an Uncertainty-Aware Adaptive Fair classification tree algorithm designed to mitigate discrimination. 
Our approach involves integrating a penalty into the Information Gain of the traditional CART algorithm. This penalty is linked to the confidence interval of the fairness metric for each potential split, taking into account both the level of discrimination and the associated uncertainty.

In constructing classification trees like CART, the Gini index is a commonly used impurity measure to evaluate potential splits at each node. For a node $ v$ containing a subset of $n_v$ units, the Gini impurity measure in this binary case is defined as
$$G_v = 2 \, \widehat{p}_{1v}(1-\widehat{p}_{1v})$$
where $\widehat{p}_{1v}$ is the observed proportion of units in node $v$ with $Y_i =1$. When considering a potential split of the node $v$ on a predictor $X_j$, the sample space $\mathcal{X}_j$ is partitioned into two regions based on a threshold $t$. Units in $v$ such that $X_{ij} \leq t$,  are assigned to the left child node, while those with $X_{ij} > t$ go to the right child node. The Information Gain from a split is calculated as the reduction in Gini impurity due to the split, given by
$$
IG(v,j,t) = G_{v} - \left ( \frac{n_{\text{left}}}{n_v} G_{\text{left}} 
 + \frac{n_{\text{right}}}{n_v} G_{\text{right}} \right ),
$$
where $ G_{\text{left}} $ and $ G_{\text{rigth}} $  are the Gini impurity measures of the left and right nodes,  each with cardinality $n_{\text{left}}$ and $n_{\text{right}}$, respectively.

At each node, the CART algorithm evaluates all possible splits on all features $\mathbf{X}_i$ and selects the one that maximizes the Information Gain. This process recursively partitions the data to find child nodes that are as pure as possible with respect to the response variable $Y_i$. The splitting continues until a stopping criterion is met, such as a minimum number of units in a node or a maximum tree depth. 

To incorporate fairness into the construction of the tree, we utilize the confidence interval CI$(\Delta_v)$ in \eqref{eq:CI} for Statistical Parity, computed on the units in $v$ and the predictions after the split on $X_j$ with threshold $t$. The Information Gain is then modified as 

\begin{equation}
    IG_{\text{FAIR}}(v,j,t) = \begin{cases} 
    IG(v,j,t) & \text{if } 0 \in \text{CI}{ (\Delta_v)} \\ 
    \lambda \, (1-\phi_{\text{CI}{( \Delta_v)}})\, IG(v,j,t) & \text{if } 0 \notin \text{CI}{(\Delta_v)} 
    \end{cases}
\end{equation}
where $ \lambda \in [0,1] $ is a tuning parameter that controls the trade-off between accuracy and fairness, and $\phi_{\text{CI}(\Delta_v)}$ is the minimum distance of the confidence interval extremes from zero, and, for intervals not overlapping zero,  it can assume values from zero to one. This quantity represents how far the predictions after the split are from fairness, accounting for the uncertainty in estimating $\Delta$. The larger $\phi_{\text{CI}{(\Delta_v)}}$ is, the more the Information Gain of node $v$ is penalized, effectively decreasing the chance that the split is selected by the algorithm. 
Therefore, for a given $\lambda$, the penalty factor adapts to the amount of discrimination involved and its uncertainty. If the predictions are fair, no penalty is added, and the algorithm proceeds as an ordinary CART. The proposed approach is outlined in Algorithm \ref{alg:fair_loss}. 

\medskip
\begin{algorithm}[ht]
\caption{Uncertainty-Aware Adaptive Fair-CART Algorithm}
\label{alg:fair_loss}
\begin{algorithmic}[1]
\Require Dataset $D$, Sensitive attribute $S$, Response variable $Y$, Tuning parameter $\lambda$
\State Initialize an empty decision tree $T$ with a root node
\State Build the decision tree $T$ using CART algorithm using $IG_{\text{FAIR}}$ where
\For{each potential split of a node $v$ in $T$}
    \State Compute the Information Gain $IG(v,j,t)$ 
    \State Compute the fairness metric $\Delta_v$ and the confidence interval CI$(\Delta_v)$
    \If{CI$(\Delta_v)$ does not include zero}
        \State Compute  $\phi_{\text{CI}(\Delta_v)} = \min \left\{ |\text{CI}(\Delta_v)_{\text{lower}}|, |\text{CI}
        (\Delta_v)_{\text{upper}}| \right\}$
        \State Compute the penalized Information Gain  $IG_{\text{FAIR}} = \lambda \, [1 - \phi_{\text{CI}(\Delta_v)} ] \times IG(v,j,t)$
        \Else
                \State $IG_{\text{FAIR}} = IG(v,j,t)$
    \EndIf
\EndFor
\State \textbf{Output:} Decision tree $T$ with fairness adapted Information Gain
\end{algorithmic}
\end{algorithm}

\medskip

It is important to note that the mitigation of discrimination  has the potential to result in a decreased accuracy.
According to \cite{breiman1984classification}, under certain regularity conditions, the CART algorithm is consistent; as the sample size increases, the sequence of classification trees converges to the Bayes optimal classifier. However, inducing fairness may affect the consistency of the algorithm. Specifically, when the underlying population is not fair, our modified algorithm may no longer be consistent in the traditional sense, as it prioritizes fairness alongside accuracy.
A consequence of this drawback, known in the literature as the \textit{impossibility theorem} \citep[see, for instance,][]{zhao2022inherent} is that our algorithm tends to overfit less than the ordinary CART algorithm with the same depth. The incorporation of the fairness penalty effectively acts as a regularization mechanism, discouraging complex splits that may lead to overfitting. As a result, instead of performing post-hoc pruning, we set the maximum depth of our tree to match the optimally pruned tree depth determined when using the ordinary CART algorithm. This approach helps balance the trade-off between model complexity, accuracy, and fairness.

Finally, note that no prior setting of a threshold for SP is required. Instead,  the trade-off between accuracy and fairness is governed by the parameter $\lambda$. This parameter can be set a priori on the based on subject-matter considerations or learned from data.

To choose $\lambda$ from data, we suggest splitting the training data into two subsets. The first subset is used to train fair trees over a grid of $\lambda$ values ranging from $0$ to $1$.
The second subset is used as a validation set to measure the predicting performance and the fairness metric for each $\lambda$. By plotting a trade-off curve, you can visually represent how different $\lambda$ values affect the two objectives---accuracy and fairness---allowing for an informed choice of this important parameter. 
Once $\lambda$ is chosen, the final tree is trained on the complete training set. A separate test set can be used for an honest estimate of the prediction error and the fairness metric. We outline all these steps in Algorithm \ref{alg:lambda}.

\medskip
\begin{algorithm}[ht]
\caption{Uncertainty-Aware Adaptive Fair-CART Algorithm with Data-Driven $\lambda$ Selection}
\label{alg:lambda}
\begin{algorithmic}[1]
\Require Dataset $D$, Sensitive attribute $S$, Response variable $Y$
        \State \textbf{Random Splitting:} Split $D$ into a training set $D_{\text{train}}$, validation set $D_{\text{val}}$, and test set $D_{\text{test}}$.
        \State \textbf{Define $\lambda$ Grid:} Define a grid of $\lambda$ values, $\{ \lambda_1, \lambda_2, \dots, \lambda_K \}$, where $\lambda_i \in [0,1]$ for $i = 1, \ldots, K$.
        \For{each $\lambda_i$, $i = 1, \ldots, K$}
         \State \textbf{Train Model:} Train Uncert.-Aware Adapt.\ Fair-CART model $T_i$ on $D_{\text{train}}$ using $\lambda_i$.
        \State \textbf{Evaluate Model:} Calculate accuracy $A_i$ and fairness metric $\Delta_i$ on $D_{\text{val}}$.
    \EndFor
    \State \textbf{Analyze Trade-offs:} Create a trade-off curve of accuracy vs fairness, with $\{A_i, \Delta_i\}_{i = 1}^K$.
    \State\textbf{Select $\lambda$:} Choose $\lambda^*$ such that the fairness metric improves significantly with minimal loss in accuracy.
       \State \textbf{Final Model Training:} Retrain the Uncertainty-Aware Adaptive Fair-CART model on the combined dataset $D_{\text{train}} \cup D_{\text{val}}$ using $\lambda^*$.
        \State \textbf{Final Evaluation:} Evaluate the final model on $D_{\text{test}}$ to obtain unbiased estimates of accuracy and fairness.
      \State \textbf{Output:} Final Uncertainty-Aware Adaptive Fair-CART model with selected $\lambda^*$.   
    \end{algorithmic}
\end{algorithm}

\begin{figure}[t]
\centering\includegraphics[width=12cm]{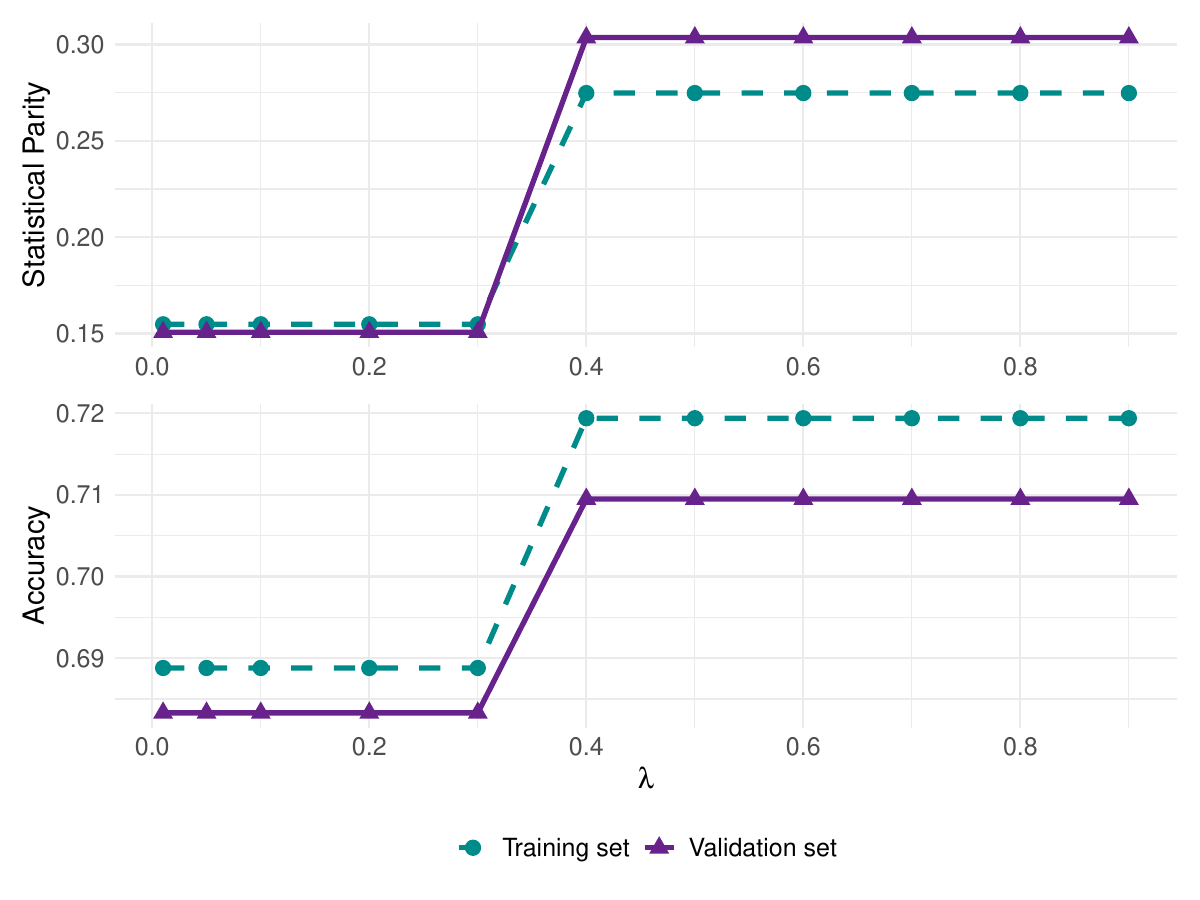}
\caption{Relationship between the $\lambda$ parameter and the two evaluation metrics, Statistical Parity and Accuracy, for the Syntethic Dataset.}
\label{fig:lambda_2k}
\end{figure}

\medskip
As an example, Figure~\ref{fig:lambda_2k} shows the trade-off curves computed for the synthetic data described in Section \ref{sec:datasets}. The first graph depicts SP curves, while the second graph illustrates the accuracy curves, on both the training and validation sets. These plots demonstrate that, in general, higher accuracy is associated with greater discrimination.  The metrics often exhibit piecewise constant behavior, reflecting the discrete changes in the tree structure as $\lambda$ varies. This indicates that the algorithm maintains consistent level of fairness and accuracy when $\lambda$ is in certain ranges, which is a positive sign for model reliability.  
For instance, in Figure~\ref{fig:lambda_2k}, users can choose between two combination of fairness and accuracy. If fairness is a priority, selecting a smaller  $\lambda$ value, such as $0.1$, is advisable. Conversely, if accuracy is more important, choosing a higher $\lambda$, like $0.4$, may be preferable.
 
The algorithm has been implemented in the R package \texttt{AdaFairTree} and it will be available at the GitHub repository \url{https://github.com/agottard/AdaFairTree}.

%

\section{Illustrative examples}\label{sec:exper}

\subsection{Datasets}
\label{sec:datasets}
To evaluate the performance of the proposed Uncertainty-Aware Adaptive Fair-CART algorithm, we consider both a synthetic dataset and four publicly available benchmark datasets. The four real-world datasets selected are COMPAS, Adult, German, and Student. These datasets are widely used in fairness research due to their inclusion of sensitive attributes such as race, gender, and age. These datasets have been shown to exhibit potential biases when applied to standard machine learning algorithms, making them ideal for evaluating fairness-aware approaches. 

\begin{figure}[h]
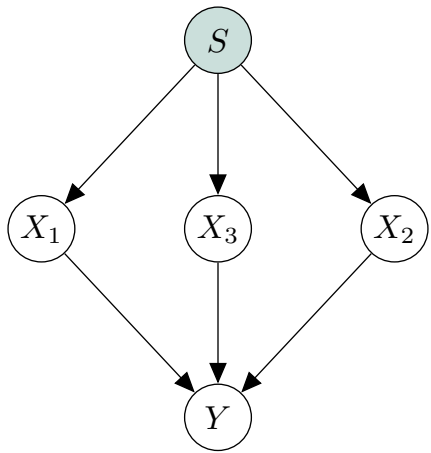

\centering
 \scalebox{1.25}{\tikz{
    \node[latent, fill=mygreen] (1) {$S$};
    \node[latent, left=1.5, yshift=-2cm] (2) {$X_1$};
    \node[latent, right=1.5, yshift=-2cm] (3) {$X_2$};
    \node[latent,  yshift=-2cm] (4) {$X_3$};
    \node[latent, yshift=-4cm] (5) {$Y$};
    \edge {1}{2,3,4};
    \edge {2,3,4} {5}
    }}
\caption{Data generating process for the synthetic data}
\label{dag}     
\end{figure}

\medskip
\noindent\textbf{Synthetic Dataset.} We adopt the data-generating process proposed by \cite{gottard2020note}, whose directed acyclic graph is shown in Figure \ref{dag}. 
This data generating process provides a typical example of data where the tree-based algorithms that use a greedy search approach, may prioritize the background variable over the direct predictors, potentially introducing additional discrimination.
The synthetic dataset has size $2,000$ and it is randomly generated by assuming that $S$, the background variable that we identify as the sensitive attribute, follows a Bernoulli distribution, $\text{Ber}(0.5)$. Each of the direct predictors $X_1, X_2$ and $X_3$ is generated from a Gaussian distribution with a mean of $(-1 + 1.1\cdot S)$ and a variance of $1$. The response variable $Y$ is generated from a Bernoulli distribution  with parameter computed using the inverse-logit function of $\eta = -1 + 0.8\cdot X_1^2 + 0.8 \cdot \sin(X_2+X_3)$.
This data generating process implies that the response  $Y$ is directly influenced by three predictors $X_1, X_2$ and $X_3$ and only indirectly affected by the background variable $S$, with $Y \ind S \mid X_1, X_2, X_3$.
\medskip

\noindent\textbf{COMPAS Recidivism.} 
The COMPAS (Correctional Offender Management Profiling for Alternative Sanctions) dataset contains information about criminal defendants in Broward County, Florida \citep{angwin2022machine}. Released by ProPublica in 2016, the dataset includes data on individuals arrested between January 2013 to December 2014, and includes variables such as age, gender, race, prior criminal history, and the COMPAS risk score. The response variable indicates whether an individual reoffended within two years. ProPublica's analysis of the COMPAS dataset revealed racial and age biases in the risk scores. Specifically, the algorithm tended to overestimate the recidivism risk for African American defendants compared to Caucasian defendants with similar criminal histories. 
\medskip

\noindent\textbf{Adult Income.}
The Adult Income dataset contains  information from the 1994 U.S. Census database \citep{kohavi1996adult}. It includes $48,842$ observations with $13$ attributes such as age, work class, education, marital status, occupation, race, gender, and native country. The response variable is a binary classification of whether an individual's annual income exceeds $\$ 50,000$. This dataset is frequently used to study fairness in income prediction models, particularly potential biases related to gender and race.
\medskip

\noindent\textbf{German Credit.}
The German Credit dataset contains information about bank account holders and their credit risk assessment \citep{groemping2019south}. It includes $1,000$ observations with $22$ predictors such as age, gender, job, housing, savings, checking account status, credit history, and loan purpose. The target variable is a binary classification indicating whether the credit risk is considered good or bad. This dataset is often used to study fairness in credit scoring models, particularly with respect to age and gender. Notably,  individuals younger than 25 seem to be discriminated in credit evaluation. 
\medskip

\noindent\textbf{Student Performance.}
The Student Performance dataset \citep{cortez2008student} provides detailed information on students' academic outcomes in Portuguese language courses. It includes $649$ students and $33$ attributes such as gender, age, parental education, study time, absences, family support, previous grades, and behavioral aspects like alcohol consumption. The response variable is the students' final grade, $G3$, which we convert into a binary outcome assuming value \textit{pass} if $G3 \geq 10$, and \textit{fail} otherwise. The sensitive attribute in this dataset is gender.

\bigskip 

The main characteristics of the datasets are summarized in Table \ref{tab:0}.

\begin{table}[H]
\caption{Key characteristics of the datasets after pre-processing}
\label{tab:0}
\centering
\resizebox*{0.9\textwidth}{!}{\renewcommand\arraystretch{1.5} \begin{tabular}{@{}llrcll}
\hline
 \textbf{Dataset}  &\textbf{Domain}& \textbf{Size} & \textbf{Variables}&  \textbf{Sensitive Variable} 
&\textbf{Target}\\
\hline
Synthetic  &-& 2000 & 5&  S 
&$Y $\\
COMPAS  &Law& 14797& 15&  Race, Gender 
&High Risk \\
Adult &Finance& 32481& 13&  Gender &Income $>$50K \\
German &Finance& 1000 & 22&  Age &Good Customer \\
 Student & Education& 649 & 33& Gender &Pass Portuguese Exam\\
 \hline
\end{tabular}}
\end{table}


\subsection{Results }

The model's performance is evaluated using two key metrics: accuracy, as the classification error rate, and SP. 
To assess the level of discrimination present in the population, we compute the SP with respect to $S$ and the observed values of $Y$, on each dataset. This measure serves as an estimate of SP in the population. 
We then compare the performance of Fair-CART with that of the standard CART algorithm, and evaluate both models with respect to the population fairness. 

Following the holdout approach, each dataset is divided into two subsets using a $70/30$ ratio, forming a training and a test set. These sets are used to train and evaluate the standard CART algorithm.
The training set is further split into a training and a validation set, again using a using a $70/30$ ratio, to select $\lambda$ from data for the proposed Uncertainty-Aware Adaptive Fair-CART algorithm, following the Algorithm \ref{alg:lambda}.
For simplicity, in this section the proposed Uncertainty-Aware Adaptive Fair-CART algorithm is referred as Fair-CART.

All results are shown in details in Table \ref{tab:results}. 
Across all datasets, the Fair-CART consistently demonstrates the ability to reduce discrimination. 

For the Synthetic data and COMPAS dataset with both race and gender as sensitive attributes, CART substantially increases the discrimination level in its predictions. In contrast, Fair-CART reduces it to a level even lower than the estimated population bias. Moreover, in both cases,  Fair-CART maintains a commendable level of accuracy. For illustration, we report in the Appendix the plots of decision trees built with the CART and the Fair-CART algorithms.

For the Adult dataset, although CART already improves fairness compared to the overall population, Fair-CART enhances it even further, achieving a SP value of 0.139 in the test set. Notably, this improvement in fairness does not come at the cost of accuracy, as the test accuracy remains high at 0.848. This highlights how the fairness-accuracy trade-off can be effectively managed by the proposed algorithm. 
Furthermore, when comparing our results on this specific dataset with those from other studies, our method outperforms all the others by achieving the lowest discrimination while maintaining the highest accuracy. Specifically, as reported by \cite{castelnovo2022fftree}, \cite{kamiran2010discrimination} achieved a SP of 0.226 with an accuracy of 0.839, \cite{zhang2019faht} reached a SP of 0.163 with an accuracy of 0.818, \cite{aghaei2019learning} obtained a SP of 0.060 alongside an accuracy of 0.815, whereas \cite{castelnovo2022fftree} reached a SP of 0.049 at their best, but with a slightly lower accuracy of 0.801. This superior performance is due to the ability of our algorithm to mitigate fairness concerns without removing or restricting specific splits in the decision process, allowing for better preservation of the model's predictive power.

 \begin{table}[H]
    \centering
    \caption{Comparison of accuracy and fairness measures for training and test sets across all datasets (Alg. = Algorithm, $d$ = Tree depth, Acc. = Accuracy, Fair-CART = Uncertainty-Aware Adaptive Fair-CART). In the \textit{sample}, metrics are computed using the observed values of the response variable $Y$, using the entire dataset. For both algorithms, metrics are calculated based on the predicted values in the training and test sets.}
    \label{tab:results}
   \resizebox*{0.9\textwidth}{!}{\renewcommand\arraystretch{1.5} \begin{tabular}{llccccccccc}
        \toprule
        \textbf{Data} & \textbf{Alg.} &  & & \multicolumn{3}{c}{\textbf{Metrics - Training}} & \multicolumn{3}{c}{\textbf{Metrics - Test}} \\ \cmidrule(lr){5-7} \cmidrule(lr){8-10}
        & & $d$ & $\lambda$  & \textbf{$\Delta$} & \textbf{$CI(\Delta)$} & Acc. & \textbf{$\Delta$} & \textbf{$CI(\Delta)$} & Acc. \\ \midrule
        \multirow{3}{*}{Synthetic} 
                                   & sample & - & - & 0.164 & (0.120, 0.207) & - & 0.164 & (0.120, 0.207) & - \\
                                    & CART  & 3 & - & 0.285 & (0.235, 0.335) & 0.717 & 0.280 & (0.203, 0.356) & 0.707 \\
                                    & Fair-CART  & 3 & 0.1 & 0.152 & (0.106, 0.199) & 0.689 & 0.135 & (0.064, 0.207) & 0.672 \\ 
                                    
        \midrule
        
        \multirow{3}{*}{COMPAS$^1$} 
                                      & sample & - & - & 0.116 & (-0.133, -0.099) & - & 0.116 & (-0.133, -0.099) & - \\
                                      & CART& 10 & - & 0.231 & (-0.250, -0.212) & 0.709 & 0.240 & (-0.269, -0.211) & 0.691 \\
                                      & Fair-CART & 10 & 0.005 & 0.041 & (-0.053, -0.028) & 0.588 & 0.061 & (-0.079, -0.043) & 0.606 \\ \midrule
        \multirow{3}{*}{COMPAS$^2$} 
                                      & sample & - & - & 0.135 & (0.115, 0.155) & - & 0.135 & (0.115, 0.155) & - \\
                                      & CART& 11 & - & 0.141 & (0.117, 0.165) & 0.707 & 0.172 & (0.136, 0.208) & 0.691 \\
                                      & Fair-CART & 11 & 0.01 & 0.062 & (0.037, 0.086) & 0.684 & 0.062 & (0.025, 0.099) & 0.656 \\ \midrule
        \multirow{3}{*}{Adult} 
                                      & sample & - & - & 0.196 & (0.188, 0.205) & - &  0.196 & (0.188, 0.205) & - \\
                                      & CART & 7 & - & 0.159 & (0.150, 0.167) & 0.846 & 0.151 & (0.137, 0.164) & 0.841  \\                        & Fair-CART  & 7 & 0.5 & 0.142 & (0.134, 0.151) & 0.854 & 0.139 & (0.126, 0.152) & 0.848 \\ 
        \midrule
        \multirow{3}{*}{German} 
                                      & sample & - & - & 0.149 & (0.073, 0.226) & - & 0.149 & (0.073, 0.226) & - \\
                                      & CART& 3 & - & 0.052 & (-0.032, 0.137)$^*$ & 0.751 &0.105 & (-0.015, 0.226)$^*$ & 0.697\\
                                      & Fair-CART & 3 & 0.05 & 0.000 & (-0.065, 0.066)$^*$ & 0.764 & 0.012 & (-0.105, 0.082)$^*$ & 0.723 \\   
        
        \midrule
        \multirow{3}{*}{Student} 
                                      & sample & - & - & 0.057 & (-0.115,  0.000)$^*$ & - & 0.057 & (-0.115,  0.000)$^*$ & - \\
                                      & CART& 3 & - & 0.021 & (-0.087,  0.045)$^*$ & 0.952 &0.051 & (-0.151,  0.049)$^*$ & 0.923\\
                                      & Fair-CART & 3 & 0.1 & 0.047 & (-0.119,  0.026)$^*$ & 0.941 & 0.039 & (-0.144,  0.066)$^*$ & 0.912 \\

        \bottomrule
        \multicolumn{6}{l}{\footnotesize{$^1$Sensitive Variable: Race (0 = Afro-American, 1 = Caucasian)}}\\
        \multicolumn{6}{l}{\footnotesize{$^2$Sensitive Variable: Gender (0 = Female, 1 = Male)}}\\
        \multicolumn{6}{l}{\footnotesize{$^*$The CI contains zero; the model's predictions are considered fair.}} 
    \end{tabular}}
\end{table}

 As for the German dataset, the both CART and Fair-CART can reduce the SP to nearly zero, with a confidence interval which includes zero. In particular, Fair-CART achieves zero SP with more accurate predictions than CART.
 
 The Student dataset provides a clear example demonstrating that when there is little to no discrimination present in the sample, the Fair-CART  model performs just as well as the standard CART model. In the original sample, the value of SP indicates no significant discrimination and this lack of bias is reflected in both the CART and Fair-CART models, where the SP remains close to zero. The CART and the Fair-CART model show comparable performance, effectively maintaining fairness and accuracy even when no fairness intervention is necessary. This result supports the robustness of the Fair-CART approach in handling datasets where discrimination is not a concern.


\section{Final remarks}\label{sec:conclusions} 
This article introduces the Uncertainty-Aware Adaptive Fair-CART algorithm, which integrates fairness assessment directly into the classification tree learning process by utilizing an uncertainty-aware fairness metric. 
This approach addresses the ongoing challenge of bias in machine learning models, particularly in classification trees. The key feature of the proposed algorithm is to penalize the information gain of discriminating splits taking into account of the uncertainty in the fairness metric via its confidence interval.
Unlike existing methods, our penalization is statistically reliable, adaptive and does not depends on pre-fixed thresholds. Instead, it adapts to the data through a tuning parameter $\lambda$, allowing for flexible balancing between fairness and accuracy based on the specific needs of each application. Such adaptability supports more informed decision-making processes when reconciling fairness and predictive performance in real-world scenarios.

The efficacy of the proposed algorithm was tested on both synthetic and benchmark datasets, demonstrating that Uncertainty-Aware Adaptive Fair-CART effectively reduces discriminatory outcomes with minimal impact on accuracy. In several cases, the model significantly reduced discrimination, measured by statistical parity, while maintaining competitive accuracy compared to standard CART algorithm. However, as expected, there is often a trade-off between fairness and performance, with slight reductions in accuracy observed in some datasets when fairness is prioritized.

Although Uncertainty-Aware Adaptive Fair-CART is a purely in-processing method, it can be extended with a post-processing step, such as relabeling at the terminal nodes, as demonstrated by \cite{kamiran2010discrimination}. This extension could further refine predictions, reducing bias, though potentially at the cost of accuracy. 
Our approach can also be generalized to other fairness measures and their corresponding uncertainty evaluations, broadening its applicability to a wider range of contexts.

In conclusion, Uncertainty-Aware Adaptive Fair-CART offers a promising approach to fairness-aware modeling in domains where equitable outcomes are critical, such as criminal justice and finance. By balancing fairness and accuracy, this method contributes to mitigating biases while maintaining practical utility in decision-making systems.

\section*{Data availability statement}
 All the benchmark data are available at \url{https://archive.ics.uci.edu/}.

\bibliographystyle{chicago}  
\bibliography{biblio}

\newpage

\section*{Appendix}
\label{appendix}
\renewcommand{\thefigure}{A\arabic{figure}}
\setcounter{figure}{0} 
\begin{figure}[ht]
\centering\includegraphics[width=11.1cm]{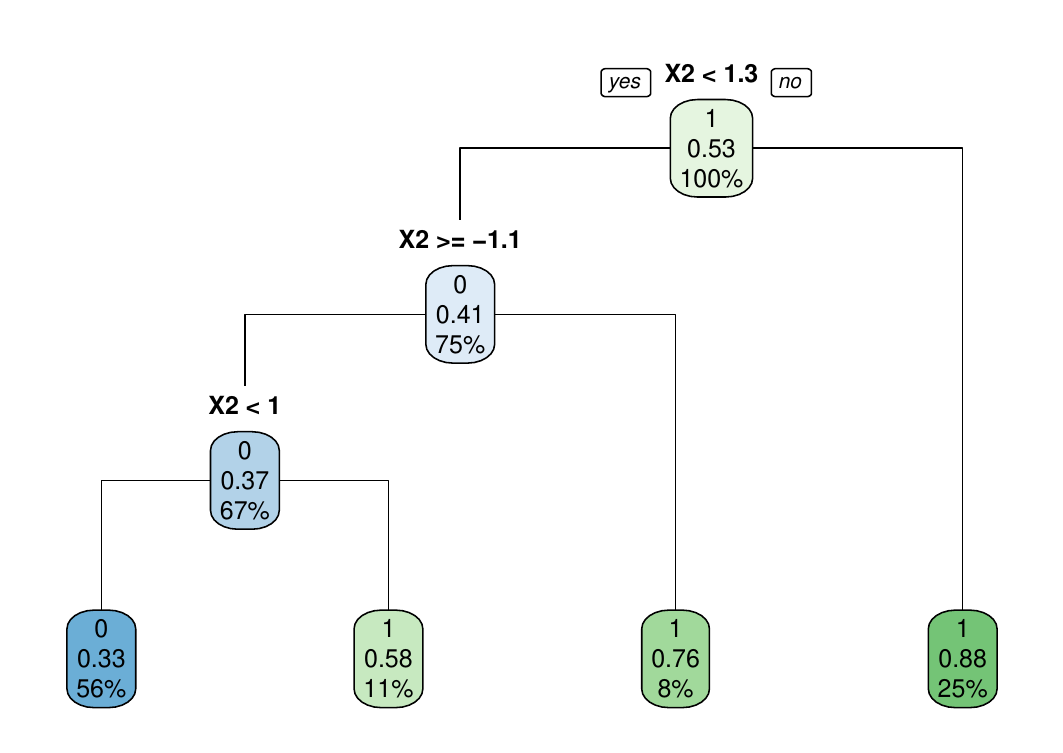}
\caption{Decision tree generated by the CART algorithm applied to the Synthetic Dataset.}
\label{fig:tree_rpart}
\end{figure}

\setcounter{figure}{1} 
\begin{figure}[ht]
\centering\includegraphics[width=11.1cm]{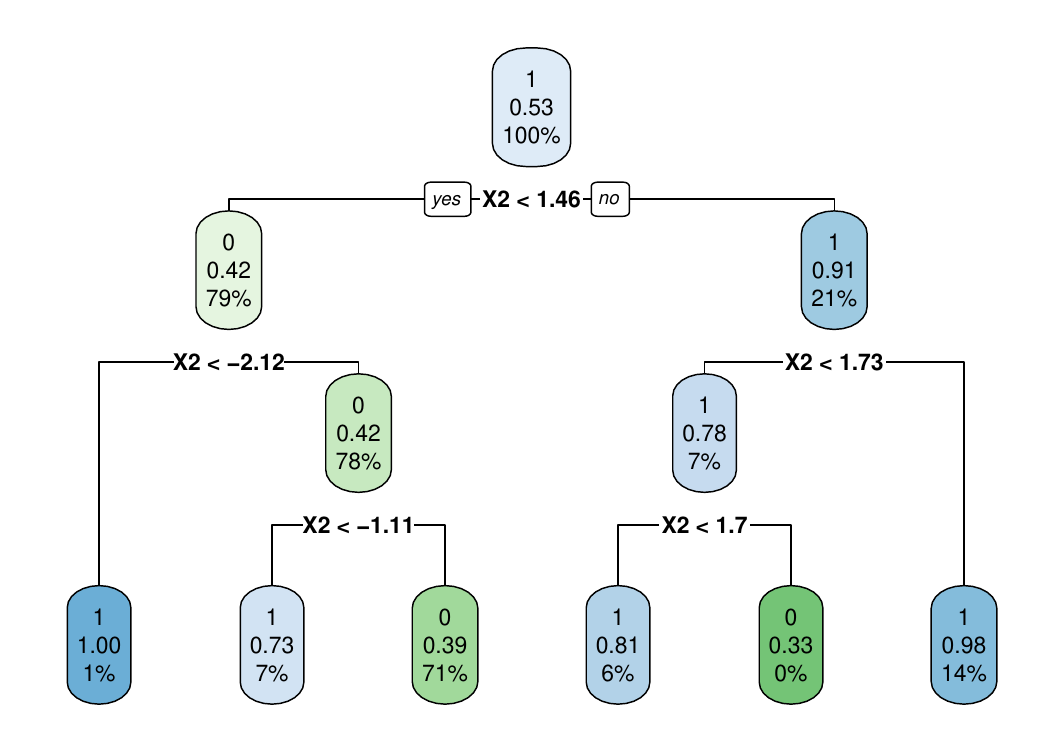}
\caption{Decision tree generated by the Uncertainty-Aware Adaptive Fair-CART algorithm applied to the Synthetic Dataset.}
\label{fig:tree_fair}
\end{figure}

\end{document}